\documentclass[10pt,twocolumn,letterpaper]{article}

\usepackage{iccv}
\usepackage{times}
\usepackage{epsfig}
\usepackage{graphicx}
\usepackage{amsmath}
\usepackage{amssymb}
\usepackage{multirow}
\usepackage{adjustbox}
\usepackage{caption}
\usepackage{multicol}
\usepackage{authblk}

\usepackage[breaklinks=true,bookmarks=false,hyperfootnotes=false]{hyperref}
\usepackage[accsupp]{axessibility}

 \iccvfinalcopy 

\def\iccvPaperID{57} 

\ificcvfinal\fi

\begin{document}

\title{\ MOSAIC: Multi-Object Segmented Arbitrary Stylization Using CLIP }
\author[1,2]{Prajwal Ganugula$^{*\dagger}$}
\author[1]{Y S S S Santosh Kumar$^{*\mathsection}$}
\author[1]{N K Sagar Reddy$^{*\ddagger}$}
\author[2]{Prabhath Chellingi$^{\curlywedge}$}
\author[1]{Avinash Thakur}
\author[1]{Neeraj Kasera}
\author[1]{C Shyam Anand}
\affil[1]{OPPO Mobiles R $\&$ D Center, Hyderabad, India}
\affil[2]{Department of Computer Science and Engineering, IIT Hyderabad, India}
\affil[ ]{\textit \url{{g.prajwal, y.kumar, nallamilli.reddy, avinash.thakur, neeraj.kasera, c.shyam.anand }@oppo.com, }   \url{cs20btech11038@iith.ac.in}}

\maketitle
\ificcvfinal\thispagestyle{empty}\fi

\begin{abstract}
Style transfer driven by text prompts paved a new path for creatively stylizing the images without collecting an actual style image. Despite having promising results, with text-driven stylization, the user has no control over the stylization. If a user wants to create an artistic image, the user requires fine control over the stylization of various entities individually in the content image, which is not addressed by the current state-of-the-art approaches. On the other hand, diffusion style transfer methods also suffer from the same issue because the regional stylization control over the stylized output is ineffective. To address this problem, We propose a new method Multi-Object Segmented Arbitrary Stylization Using CLIP (MOSAIC), that can apply styles to different objects in the image based on the context extracted from the input prompt. Text-based segmentation and stylization modules which are based on vision transformer architecture, were used to segment and stylize the objects. Our method can extend to any arbitrary objects, styles and produce high-quality images compared to the current state of art methods. To our knowledge, this is the first attempt to perform text-guided arbitrary object-wise stylization. We demonstrate the effectiveness of our approach through qualitative and quantitative analysis, showing that it can generate visually appealing stylized images with enhanced control over stylization and the ability to generalize to unseen object classes.
\let\thefootnote\relax\footnotetext{* Equal Contribution.}
\let\thefootnote\relax\footnotetext{$\dagger$ Prajwal led the project and guided the intern.}
\let\thefootnote\relax\footnotetext{$\ddagger$ Sagar provided valuable contribution to the pipeline.}
\let\thefootnote\relax\footnotetext{$\mathsection$ Santosh provided valuable contribution to text segmentation.}
\let\thefootnote\relax\footnotetext{
$\curlywedge$ Work done during internship at OPPO Mobiles R$\&$D Center.}

\end{abstract}

\section{Introduction}

Style transfer has emerged as an essential technique in the field of computer vision and image processing, allowing for the transformation of style from a style or texture image to a reference image while preserving the contents of the reference image. Gatys \etal\cite{DBLP:journals/corr/GatysEB15a} formulated style transfer as an image optimization problem, which was later implemented by Ulyanov \etal\cite{8} using a feed-forward neural network to reduce inference time. To improve the visual quality of the results, Johnson \etal\cite{4} proposed the use of perceptual loss. However, these techniques were limited to single-styling images. Dumoulin \etal\cite{dumoulin2016learned} addressed this issue by implementing Conditional Instance Normalization layers to extend the stylization network to multiple styles. However, this approach becomes infeasible after reaching a certain number of styles and is limited to the styles the model is trained on. To overcome these limitations, Xun Huang \etal\cite{huang2017arbitrary} proposed the use of Adaptive Instance Normalization to extend style transfer to arbitrary styles. Although several techniques have been proposed for style transfer, each technique has its limitations, and further research is needed to develop a more robust and flexible style transfer algorithm.

\begin{figure*}[t]
\begin{center}
\includegraphics[width=\textwidth]{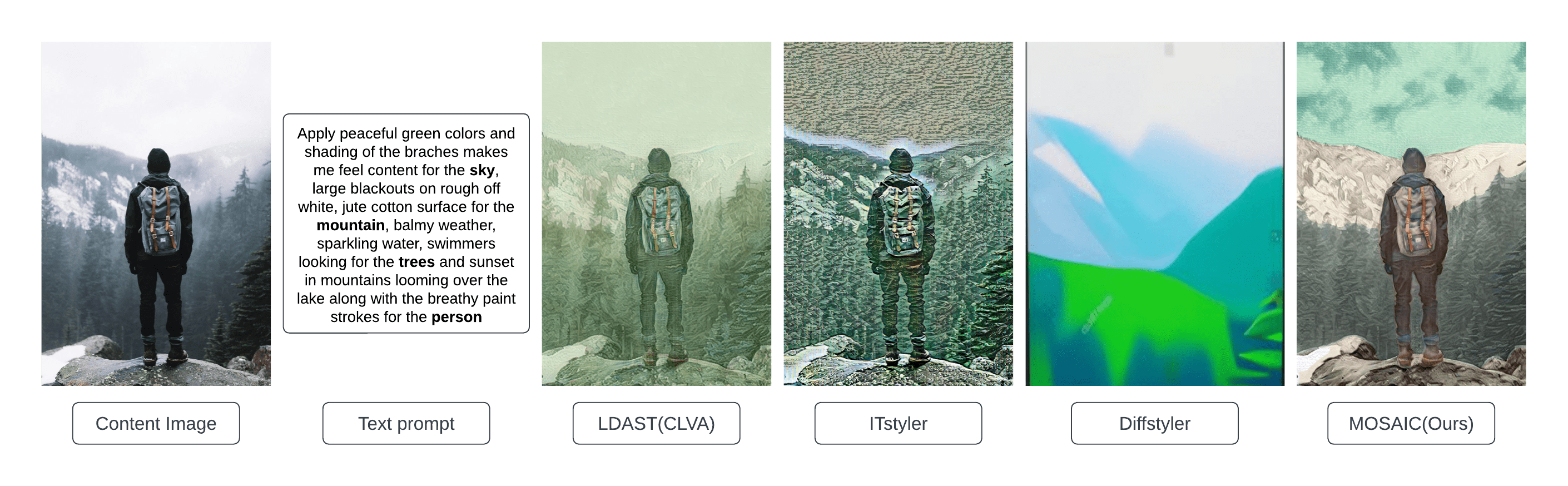}
\end{center}
   \caption{Showing Our output result comparing with \textit{LDAST}\cite{fu2021languagedriven}, \textit{ITsytler}\cite{bai2023itstyler}, \textit{Diffstyler}\cite{huang2022diffstyler},
   \textit{MOSAIC(ours)}.
   },
\label{fig: Banner}
\end{figure*}

Applying different styles to different objects in an image is a challenging problem that requires identifying individual objects and applying corresponding styles to each object. Kurzman \etal\cite{kurzman2019class} proposed a Class-Based styling method, where they utilized a segmentation model to identify the objects belonging to the same class and applied styles guided by the segmentation masks. This approach allows for consistent styling of objects in the same class and produces visually appealing results. However, this method does not consider the individual characteristics of each object, which limits its flexibility. Huang \etal\cite{10.1111:cgf.13853} proposed the Style Mixer method to address this limitation by applying multiple styles based on regional semantics. This approach allows for more precise control over the stylization of each object by considering its characteristics, leading to more diverse and creative stylization. However, this method requires additional computational resources and may lead to longer stylization times. Although both Class-Based styling and Style Mixer methods have shown promising results, further research is needed to develop more efficient and flexible techniques for object-level style transfer.



Object recognition models trained on specific object classes tend to perform poorly in recognizing unseen object classes. Additionally, objects belonging to the same class may have different descriptions within an image, requiring the model to understand illustrations of both seen and unseen classes during training. To address this limitation, Lüddecke and Ecker proposed the CLIPSeg\cite{lddecke2021image} method, which utilizes arbitrary text prompts to segment objects. Similarly, Li \etal\cite{li2022languagedriven} proposed the LSeg method for language-driven semantic segmentation, using the contrastive learning approach with text and image embeddings to predict segmentation classes. Other works like OpenSeg \cite{ghiasi2021open} and OVSeg \cite{liang2022open} have also addressed similar issues. Following these, a recent work SAM\cite{kirillov2023segment}(Segment Anything Model) was trained on a new dataset which was created mainly for segmentation. This was designed to take inputs in different ways(point based, text based, bounding box based). CLIP\cite{radford2021learning} embeddings could be passed as an input to SAM, which makes text-based Image segmentation possible.

Prompt-based arbitrary style transfer is a branch of style transfer networks where input text prompts are used instead of reference style images. Kwon et al. (2021) proposed CLIPStyler\cite{kwon2021clipstyler}, which stylizes images based on the input text descriptions of the style. They employed patch-wise CLIP(Contrastive Language-Image Pretraining)\cite{radford2021learning} loss to moderate the quality of style images in all regions. Subsequent works, including Fast CLIPStyler\cite{an2022fastclipstyler}, LDAST\cite{fu2021languagedriven}, and ITstyler\cite{bai2023itstyler}, have enhanced the ability of stylization networks. Diffusion-based models, such as DiffStyler\cite{huang2022diffstyler}, have also been used for text-guided stylization networks. Despite these advancements, there is still room for further research to develop more efficient and effective prompt-based style transfer techniques.




Existing approaches for style transfer typically require a reference style image, but no complete text-based pipeline for multi-object arbitrary style transfer currently exists. In this paper, we propose a novel text-based pipeline for multi-object arbitrary style transfer, allowing generation of stylized images with a text description of the desired style.

Our contributions can be summarized in three main points: 
\begin{enumerate}
\item We propose a novel text-based pipeline for multi-object arbitrary style transfer, which utilizes a custom decoder block to segregate text into segmentation and stylization tasks. 
\item Employing a combination of style embedding and object segmentation techniques to generate high-quality stylized images. This research has the potential to provide new and efficient ways to create stylized images without the need for reference-style ideas.
\item Conducting experiments to demonstrate the effectiveness of our approach in generating visually appealing stylized images with enhanced control over stylization. We evaluate the results using user study and patch-wise CLIP score to capture object wise stylization capability. Our model's ability to generalize to unseen object classes is shown in Figure \ref{fig: Banner}.
\end{enumerate}

\section{Related work}

\textbf{Image-based style transfer.}
Transferring styles from one image to another has garnered significant attention in recent years due to its ability to create visually appealing and artistic images. Early techniques for image-based style transfer involved pixel-wise updates of content images compared to brush-strokes of style images \cite{6243138, 790383, 1, efros2001image, heeger1995pyramid}. However, these techniques were later replaced by neural style transfer methods \cite{7780634}, \cite{DBLP:journals/corr/RuderDB16}, \cite{DBLP:journals/corr/GatysEB15a}, \cite{2}, \cite{3} that enabled the creation of more realistic stylized images. While these methods provided high-quality stylized images, the optimization process was slow. To address this issue, knowledge-distilled transform networks were introduced \cite{4}, \cite{5}, \cite{6}, \cite{7}, \cite{8} to speed up the optimization process. However, these methods were limited to single-style per-model transformation. To overcome this limitation, Conditional Instance Normalization (CIN) layers were introduced \cite{dumoulin2016learned}, \cite{li2017diversified} that enabled the creation of a multiple-style per-model network. However, CIN layers were limited to a fixed set of styles. To enable arbitrary styles per model, the CIN layers were replaced by Arbitrary Instance Normalization (AdaIN) layers \cite{huang2017arbitrary}, \cite{chen2016fast}. Further improvements were made to preserve various features of a stylized image, including AdaAttn\cite{liu2021adaattn}, AesUST\cite{wang2022aesust}, and All-to-key\cite{zhu2022alltokey} as well as various other techniques \cite{risser2017stable, peng2017synthetic, li2017demystifying}.

\textbf{Text-Based style transfer:}
Text-based style transfer has emerged as an alternative to image-based techniques due to the limitations of the latter approach in terms of the availability of style images. Language-based image editing using predefined semantic labels, referred to as LBIE, was introduced as a solution to this problem \cite{laput2013pixeltone}. Kwon proposed a text-style transfer technique using CLIP\cite{radford2021learning} in CLIPstyler\cite{kwon2021clipstyler}. However, CLIPstyler has a high inference time, leading to the development of various improvements, including FastCLIPstyler \cite{an2022fastclipstyler}, LDAST \cite{fu2021languagedriven}, and ITstyler\cite{bai2023itstyler}, which utilize normalization layers. Furthermore, significant progress has been made using Diffusion models \cite{yang2022diffusion} and GANs \cite{goodfellow2014generative} in DiffStyler \cite{huang2022diffstyler}, styleGAN \cite{karras2018stylebased}, Pix2pix \cite{isola2016imagetoimage},\cite{taigman2016unsupervised, liu2017unsupervised, arjovsky2017wasserstein, liu2016coupled, reed2016generative, oord2016pixel, kingma2013autoencoding, huang2016stacked, denton2015deep, bousmalis2016unsupervised, salimans2016improved}. To improve CLIP's performance, mixed forms with GANs were explored in styleCLIP \cite{DBLP:journals/corr/abs-2103-17249}, and NADA \cite{DBLP:journals/corr/abs-2108-00946}.

\textbf{Semantic Image segmentation:} 
Semantic image segmentation is identifying and segmenting specific objects within an image. Traditionally, large models pre-trained over extensive datasets \cite{DBLP:journals/corr/abs-2011-10566, DBLP:journals/corr/abs-2002-05709} have been used to perform this task, which is computationally expensive. Recently, neural network and transformer-based approaches have been proposed in DabNet\cite{li2019dabnet}, TransUNet\cite{DBLP:journals/corr/abs-2102-04306}, SETR\cite{DBLP:journals/corr/abs-2012-15840}, Segformer\cite{DBLP:journals/corr/abs-2105-15203}, and Segmenter\cite{strudel2021segmenter}, to improve the efficiency of the segmentation process. However, these methods have limited class segmentation capabilities. Text-based image segmentation has been proposed to segment images into infinite classes to address this limitation. Several architectures have been proposed based on CLIP, including CLIPseg\cite{lddecke2021image}, Langseg\cite{li2022languagedriven}, and CRIS\cite{wang2021cris}. Several models based on GANs\cite{goodfellow2014generative} have been proposed to enhance semantic image segmentation. The recent introduction of the SAM \cite{kirillov2023segment} model made a new revolution in image segmentation as it can segment anything. SAM is the world's first massive-scale, promptable, interactive foundation image segmentation model. For this model's training, they have gathered a new dataset of 11 million images and 1.1 billion masks. It can take input in three ways; one is prompt-based input which uses CLIP\cite{radford2021learning} to encode the prompt and mask the image. The architecture is conveniently designed to produce different masks for different inputs with faster inference once the input image gets encoded. This feature is beneficial in masking multiple objects efficiently.

\textbf{Content Based Image Style Transfer:}
Content-Based Image Style Transfer refers to stylizing each object in an image separately. CB-Styling \cite{kurzman2019class} was one of the earliest attempts in this direction, which utilized a combination of segmentation \cite{li2019dabnet} and stylization \cite{3} networks. However, a limitation of CB-Styling was its inability to handle a wide range of classes in segmentation. Several approaches have been proposed to address this limitation that use attention mechanisms to stylize objects in images by comparing them with style images. For instance, StyleMixer \cite{10.1111:cgf.13853}, Splice \cite{DBLP:journals/corr/abs-2201-00424}, MAST \cite{DBLP:journals/corr/abs-2005-10777}, and \cite{DBLP:journals/corr/abs-2108-04441} utilize attention mechanisms to stylize objects in images, although they all suffer from the drawback of being limited to the stylization of the content image in the presence of the style image, which falls short of meeting the desired quality benchmarks.

\textbf{Semantic text segmentation:}
Semantic text segmentation is a critical task in natural language processing, which involves extracting semantic features and identifying various classes from a given text. Language translation transformers \cite{vaswani2017attention} have made this process easy by connecting the words in a sentence to extract semantic meaning. Further advancements have been made using Generative Pre-trained Transformers (GPTs) \cite{brown2020language}, demonstrating an understanding of the importance and capability to perform specific tasks.

\section{Method}

\begin{figure*}
\begin{center}
\includegraphics[width=\textwidth]{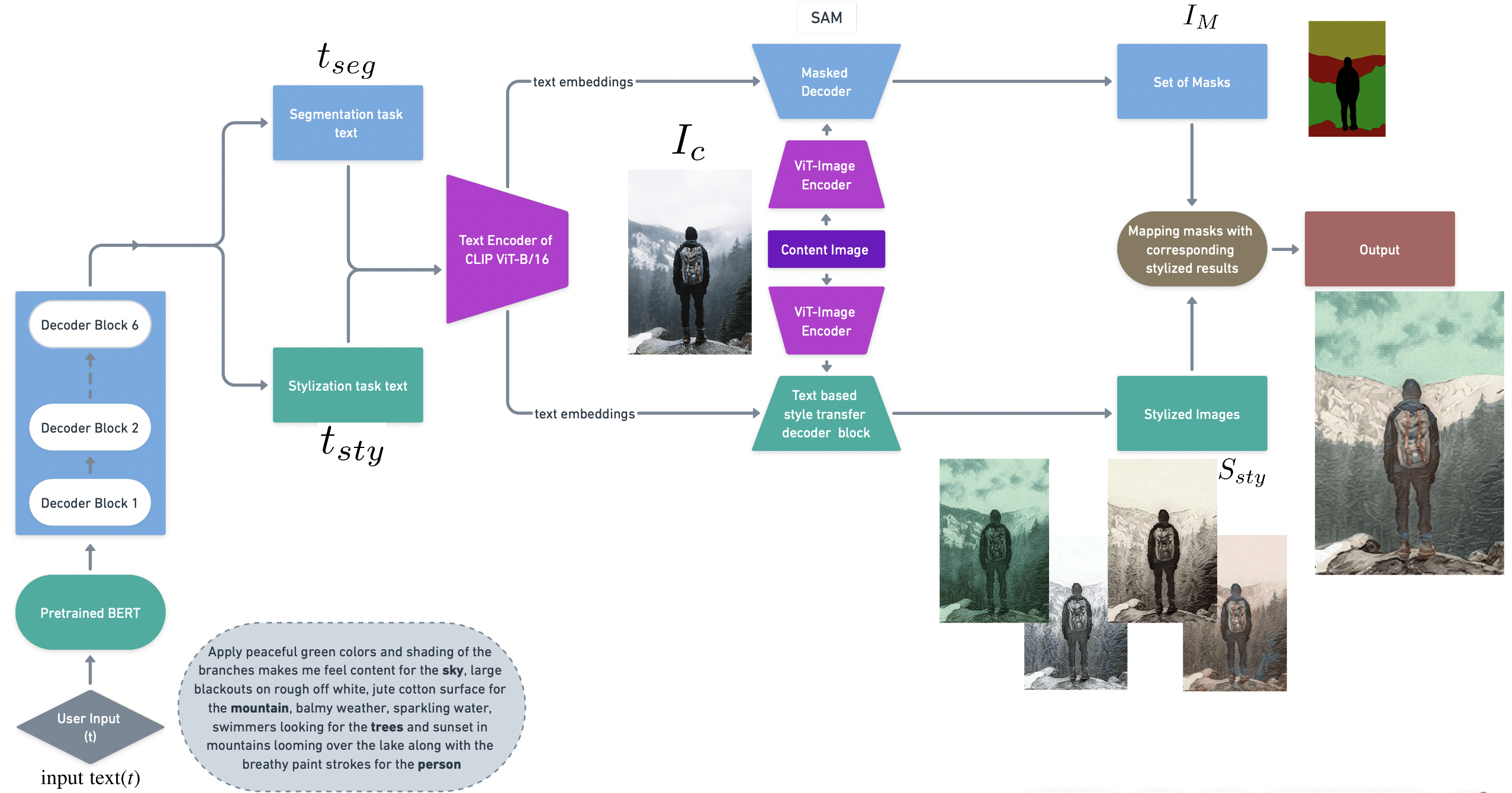}
\end{center}
   \caption{Pipeline of the proposed style transfer method showing the input flow.}
\label{fig: Architecture}
\end{figure*}

We aim to perform object-wise text-style transfer in an input image with a text prompt. To achieve this, we propose a pipeline as shown in Figure \ref{fig: Architecture}. LDAST\cite{fu2021languagedriven}(Language Driven Artistic Style Transfer) and SAM\cite{kirillov2023segment}(Segment anything Model) inspire the blocks in our pipeline. The encoders of LDAST and SAM models use pre-trained CLIP ViT-B/16 text encoder as a backbone and take embeddings from the model to further process their respective tasks. The details and architectures of these models are discussed in this section.

\begin{figure}[t]
\begin{center}
\includegraphics[width=1.0\linewidth]{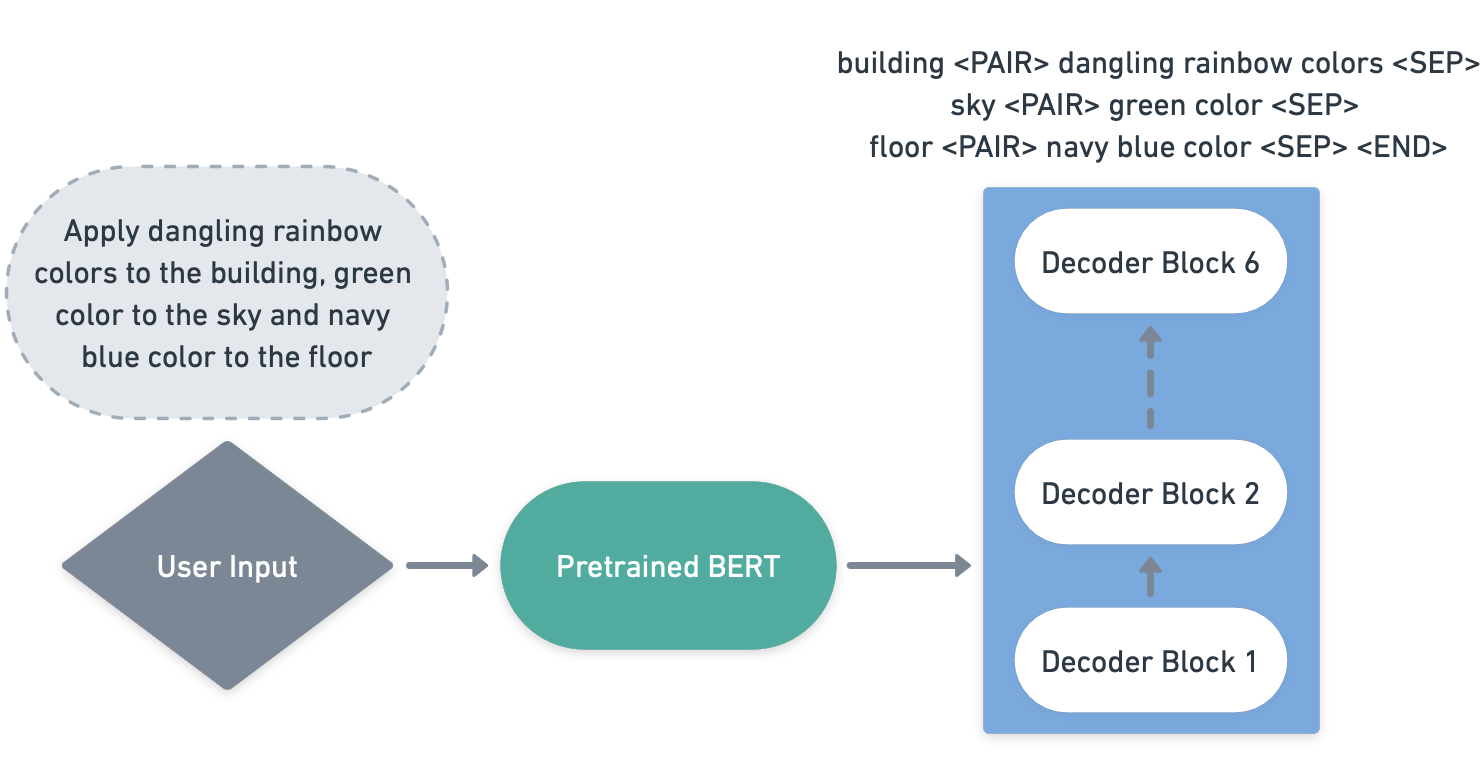}
\end{center}
   \caption{Architecture of the BERT \cite{devlin2018bert} based text segmentation model.}
\label{fig: text-segmentation-architecture}
\end{figure}

\subsection{Text Segmentation Model} \label{sec:method:text-seg-model}
Our approach's first step is segmenting the input prompt into objects along with their position description in the image and corresponding styles. We took a pre-trained BERT \cite{devlin2018bert} based model because of its strong ability to understand relationships between different segments of text (objects and their corresponding styles). We added a custom decoder to produce the objects and styles pairs. Each pair is separated by $<SEP>$ token, and the object style in every pair is separated by $<PAIR>$ token. Refer to Figure \ref{fig: text-segmentation-architecture} for the architecture. We trained our model using Text Segmentation Loss.

\subsubsection{Text Segmentation Loss} \label{sec:method:text-seg-loss}
We used cross-entropy loss \cite{lecun2015deep} for text segmentation loss $(L)$.

\begin{equation}\label{first_equation}
 L = -\sum_{c=1}y_{c}\log(p_{c})
\end{equation}

where $y_{c}$ is the ground truth and $p_{c}$ is the predicted probability of the model

We also explored the usage of GPT \cite{radford2018improving} based models for this task. Refer to [\ref{sec:exp:text-segmentation-model}] for more details.

\subsection{CLIP Based Models}
CLIP\cite{radford2021learning} is a recent initiative aimed at connecting images and text, with a particular focus on zero-shot capabilities. The CLIP model consists of a ViT \cite{dosovitskiy2020image} based image encoder and a combination of transformer blocks as text encoder. The CLIP model is trained on a large dataset of image-caption pairs using a contrastive learning approach. The model is trained to maximize the similarity between the embeddings of matching image-caption pairs, while minimizing the similarity between the embeddings of non-matching pairs. This is achieved by defining a contrastive loss function that encourages the model to learn embeddings that are discriminative for the given task. The CLIP model is useful in Joint processing of text and images. Since CLIP is contrastively trained on huge dataset, it can generalize well even for downstream zeroshot tasks. 

\begin{figure*}
\begin{center}
\includegraphics[width=1.0\textwidth]{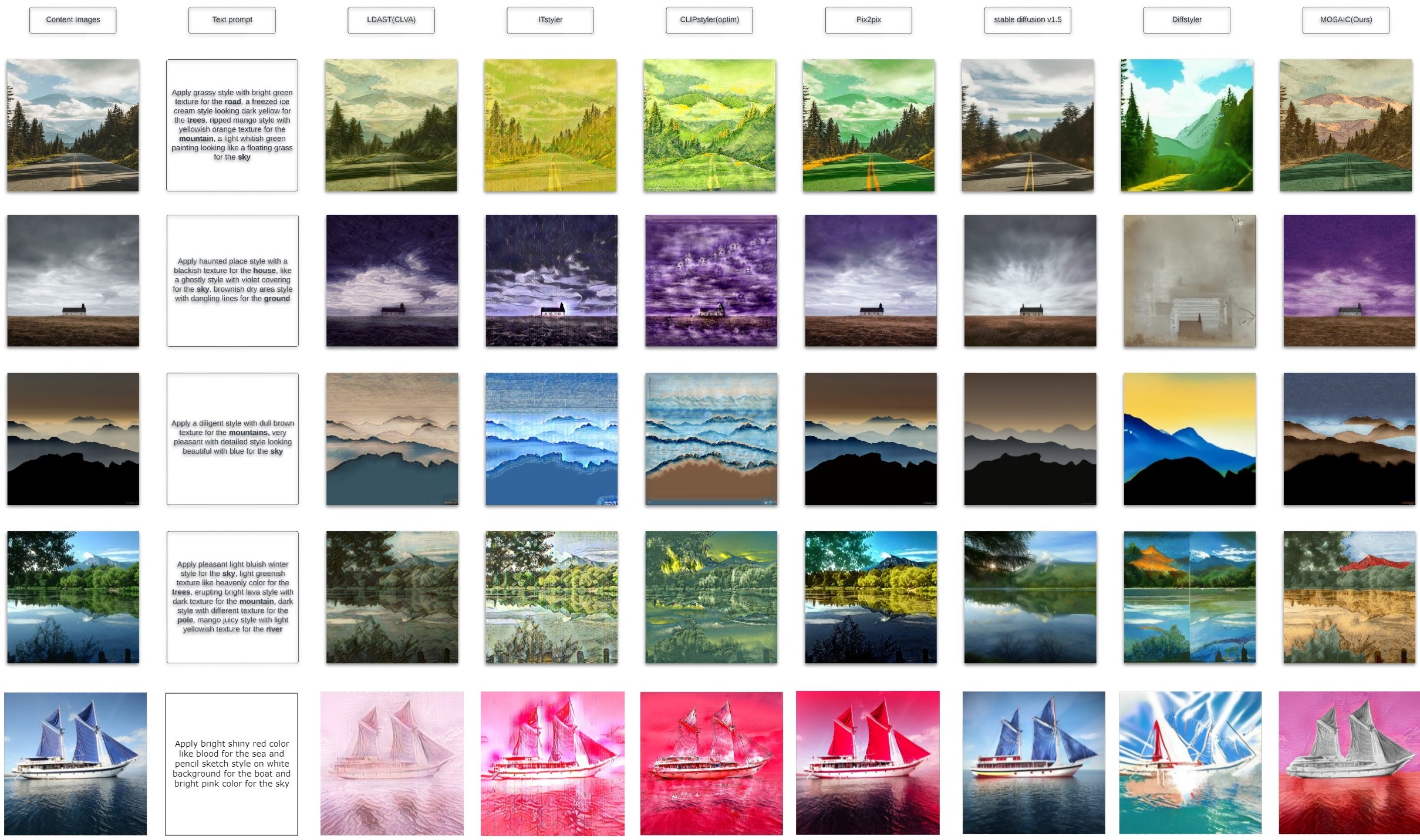}
\end{center}
   \caption{Comparing the output of our model with LDAST\cite{fu2021languagedriven}, ITstyler\cite{bai2023itstyler}, CLIPstyler(optim.)\cite{kwon2021clipstyler}, Pix2pix\cite{isola2016imagetoimage}, stablediffusionv1.4\cite{Rombach_2022_CVPR}, Diffstyler\cite{huang2022diffstyler}, MOSAIC(ours). Our model produced results that were in line with user expectations and showed superior results between stylized objects than other models.}
\label{fig: Text-based models}
\end{figure*}

\subsection{LDAST}
This model aims at text-based editing of an image. It has two modules one is LVA (Language Visual Artist), the other is CR (Contrastive Reasoning), and the combination is CLVA (Contrastive Language Visual Artist)\cite{fu2021languagedriven}. The process of extracting style from text and applying it to a content image is the core functionality of the Language Visual Artist (LVA) module within the CLVA model. The LVA module enables the network to learn to embed the style text and relate it to the corresponding style image using a discriminator. The Contrastive Reasoning (CR) module further enhances this process by comparing contrasting pairs of style images and text to improve or correct the relativeness between the outputs. To learn the art of style transfer from text, LVA employs a combination of structure reconstruction, patch-wise style discriminator, content matching, and style matching losses. The VGG\cite{simonyan2014deep} encoders and decoders are used to extract the feature maps, which are then stitched back together. The CR module improves the learned parameters by applying consistency and relativity losses, resulting in improved content consistency and the relativeness of style in the output. These advanced techniques have the potential to revolutionize text-based image editing and enable more sophisticated image stylization through text.

\subsection{Segment Anything Model (SAM)}
The objective of SAM\cite{kirillov2023segment} is to segment an image into specific classes based on the text prompt. It comes with a separately running image encoder and prompt encoder, which helps in making multiple mask generation efficient. We will cache the image encoding and reuse it whenever a new mask needs to be generated. This mask is generated from the lightweight mask decoder, which runs efficiently even on the CPU. This lightweight mask decoder takes image encoding and prompt encoding as input and produces the respective masked images. The image encoding comes from the cache, and the prompt encoding will be generated instantaneously, even over the CPU. This is all possible because of practical training of the model over the newly developed dataset\cite {kirillov2023segment}, which comprises 11 million images and 1.1 billion mask captions. This helps us effectively stylize the image object-wise efficiently.

\subsection{Architecture Pipeline}

Our Architecture pipeline in Figure \ref{fig: Architecture}, takes the above ideas to create a unified channel, seamlessly performing the required task. A content image and a text prompt are given as input which is then operated on by the modules in the pipeline in a sequential order to produce final object-wise stylized output. 
In the \textbf{text segmentation block}, the input text(\textit{t}) is segmented into stylization (\textit{$t_{sty}$}) and segmentation (\textit{$t_{seg}$}) texts. Each object word from \textit{$t_{seg}$} is mapped to corresponding style phrases in \textit{$t_{sty}$}. Each mapping from the segmented input text ($t_{sty} + t_{seg}$) is passed parallelly (or sequentially) into the next block separately. After segregating the \textit{$t_{sty}$} and \textit{$t_{seg}$}, we pass them to CLIP-ViT-B/16 pre-trained text encoder to get the corresponding text embeddings for the respective tasks. The two models(Segmentation and stylization networks) take these embeddings and process their respective tasks.
For the \textbf{image segmentation block}, we take the segmentation part of input text ($t_{seg}$) and content image ($I_c$) and produce the required object masks ($I_M$). Simultaneously at the \textbf{stylization block}, we give the $t_{sty}$ and $I_c$. This produces all the corresponding stylized images for each style in $t_{sty}$. This gives out a set ($S_{sty}$) of stylized images of $I_c$. In the \textbf{Object-wise stylization block} of the architecture, we extract the styles of objects from this $S_{sty}$ using the object masks ($I_M$) and mappings established previously in the first block. Finally, after extracting the corresponding style for each object, we combine the pixel values to produce the final MOSAIC output.

\section{Experiments}

\subsection{Text Segmentation Dataset}
To get accurate text segregation for segmentation and style transfer tasks, we need data for which the input text and corresponding segregated text segments for each of the segmentation and stylization tasks are annotated. For this reason we constructed a dataset consisting of 400 classes and 150 styles. Our dataset was carefully designed to ensure that the resulting text prompts closely resembled the inputs typically encountered in real-world scenarios. Using this dataset, we trained a text segmentation model that is capable of accurately segmenting and segregating the text prompts while preserving the relationships between the resulting segments.

\subsection{Text Segmentation Model} \label{sec:exp:text-segmentation-model}
We adopted a pretrained BERT encoder, which was integrated with our custom decoder comprising 6 decoder layers, the embedding size was set to 512, used 8 heads, as previously established in  \cite{vaswani2017attention}. To optimize the training process, we maintained the encoder's parameters fixed while exclusively training the decoder on our curated dataset, utilizing the Cross Entropy Loss (Equation \ref{first_equation}). Freezing BERT's weights resulted in fast convergence due to its ability to generalize to intricate inputs. Additionally, this led to a substantial reduction in the number of learnable parameters, enhancing training efficiency.

We used adam optimizer with initial learning rate as 0.001 and Cosine Annealing with warmup phase of 5 epochs and trained it for 400 epochs on 8 V100 GPU cluster.
Our model has showed the capability to generalize well to arbitrary text prompts which contain multiple objects and  styles. Alternatively, If we want to have a more generalized model that can generalize well to unseen classes and styles, we can also use ChatGPT API \cite{openai} to decompose the prompt text into objects and their corresponding styles.


\subsection{Qualitative Analysis}
We show the results to understand the effectiveness of our pipeline. We divided this study into two stages. Initially, we compare the results with models which come under the same reign $i.e.$, text-based style transfer models. We took some SOTA models, Pix2pix\cite{isola2016imagetoimage}, LDAST\cite{fu2021languagedriven}, ITstyler\cite{bai2023itstyler}, CLIPstyler(optim.)\cite{kwon2021clipstyler}, Diffstyler\cite{huang2022diffstyler}, and stablediffusionv1.4\cite{Rombach_2022_CVPR}, for the analysis. We excluded the FastCLIPstyler\cite{an2022fastclipstyler} and CLIPstyler(fast)\cite{kwon2021clipstyler} as the CLIPstyler(optim.) is always better than them. The results of the outputs produced with the same content and text prompts are shown in Figure \ref{fig: Text-based models}. The problem with these models is that they cannot distinguish objects from the text and its corresponding style phrases, as obtained from our proposed method.


\begin{figure*}[t]
\begin{center}
\includegraphics[width=\textwidth]{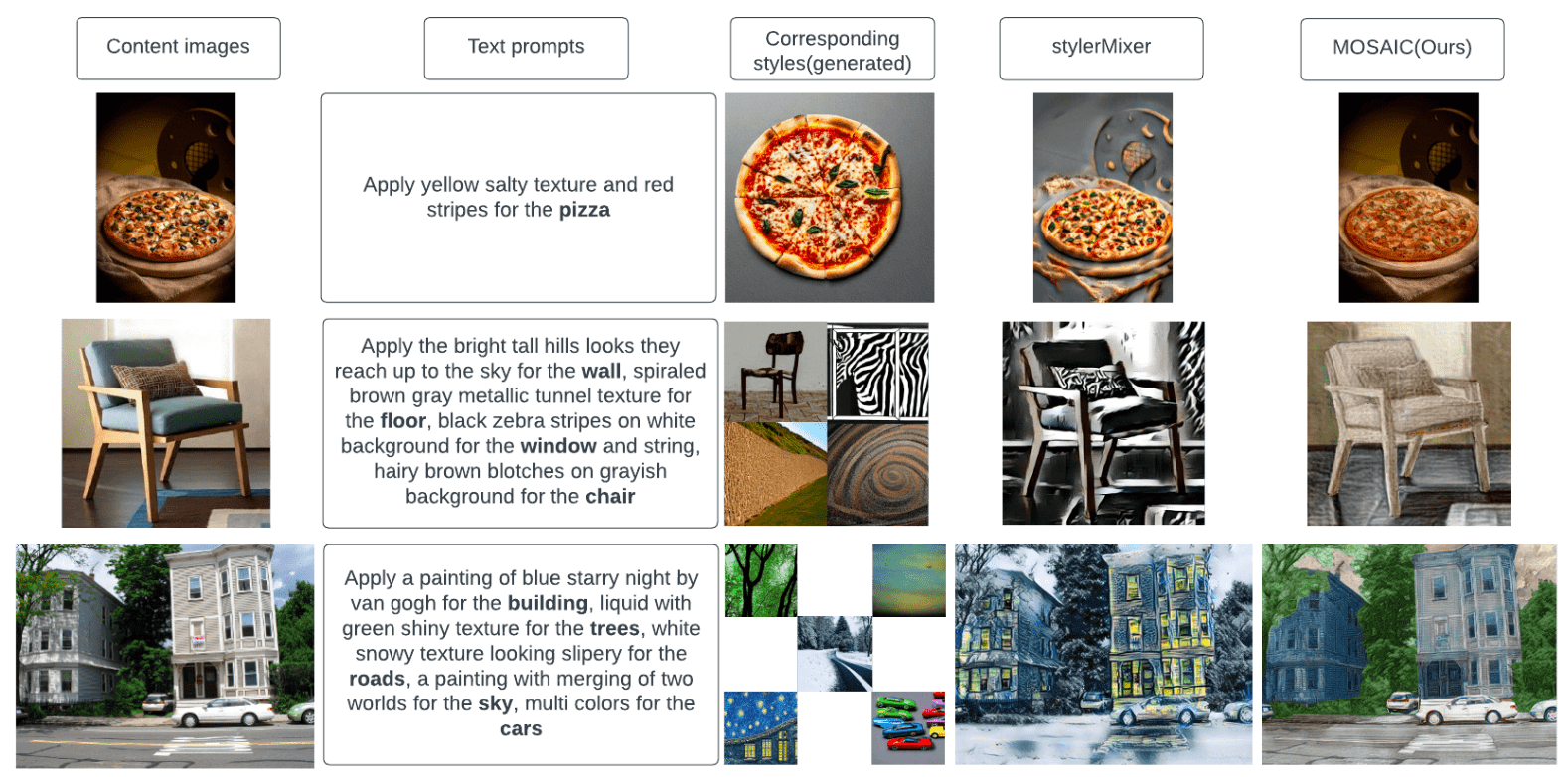}
\end{center}
      \caption{Comparing our model results with \textit{styleMixer}\cite{10.1111:cgf.13853}, showing that the styles aren't extracted well from the style images by styleMixer}  
    \label{fig: styleMixer}
\end{figure*}





The next stage includes the image-based style transfer models, which use the style images generated by the stable diffusion model of \textit{huggingface}\cite{webhuggingface}(stable-diffusion-v1.4\cite{Rombach_2022_CVPR}) from the same text prompt. We compared the results with \textit{styleMixer}\cite{10.1111:cgf.13853} by generating style images for each text in the text prompt, as shown in Figure \ref{fig: styleMixer}. The outputs of \textit{styleMixer} are not good because of two drawbacks. One is that the styleMixer needs style images with objects almost matching the content image. The other is that the stable diffusion used here can't generate the styles ideally from the same text prompt given to our model. We need to check whether the output is accurate to text rather than just seeing its uniformity. These problems don't stop our model from giving pleasing results, as it doesn't need any style image. 

\begin{figure*}
\begin{center}
\includegraphics[width=1\textwidth]{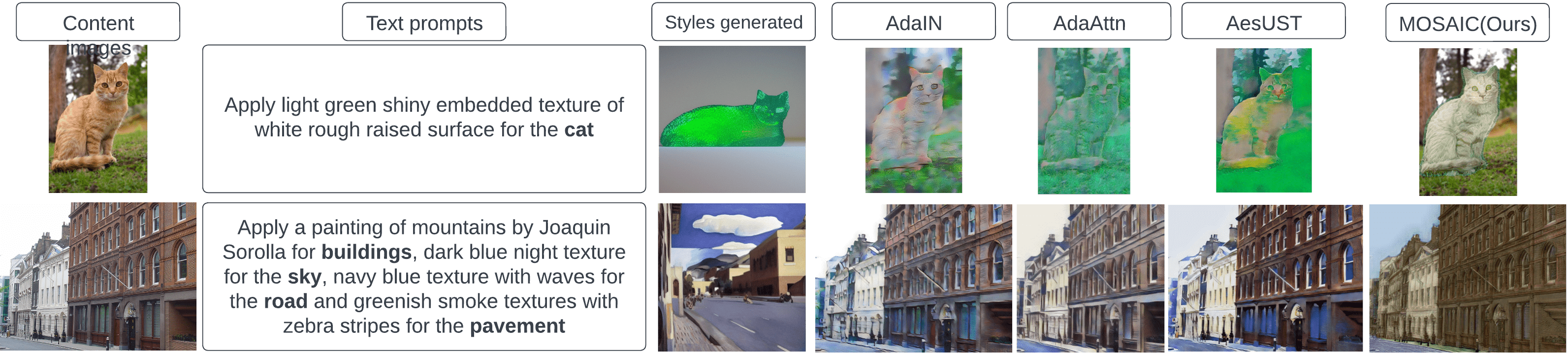}
\end{center}
   \caption{Comparing our model with SOTA Image based style transfer models, \textit{AdaIN}\cite{huang2017arbitrary}, \textit{AdaAttn}\cite{vaswani2017attention}, \textit{AesUST}\cite{wang2022aesust}, \textit{MOSAIC(ours)}. This shows that the previous models cannot perform object-specific style transfer and shows the inaccuracy of generated style image.}
\label{fig: Image-based models}
\end{figure*}

We also compare our model with Image-based style networks using the comparable style for the text prompt generated using the same stable diffusion model of \textit{huggingface}\cite{webhuggingface}(stable-diffusion-v1.4\cite{Rombach_2022_CVPR}). In examining Figure \ref{fig: Image-based models}, the results look comparable in quality, but the style transfer is not object-specific, and obtaining the style images which suit the text prompt every time can be challenging.

The advantage of models similar to ours $i.e.$, text-based models, are they don't need style images to transfer the style. Finding a style image that suits a user's description(text prompt) is difficult; hence, an image-based model will always have limitations.

\subsection{Quantitative Analysis}

\subsubsection{User Study}
For the quantitative comparison of our model, we have conducted a user study for quantitative analysis. For the user study, we have evaluated 120 Content images with 15 different stylization prompts per image for the following models: LDAST\cite{fu2021languagedriven}, ITStyler\cite{bai2023itstyler}, CLIPStyler\cite{kwon2021clipstyler}, Pix2Pix\cite{isola2016imagetoimage}, Diffstyler\cite{huang2022diffstyler} and MOSAIC. We have gathered responses from 78 users through a form, and collected their ratings for each stylized image. We have given the following scores for the users to select, from 1 being the least, to 10 being the most accurately stylized images.
The results of the User study survey can be visualized in Table \ref{RESULTS_table}.

\subsubsection{Patch-wise CLIP score}

Similar to CLIPStyler\cite{kwon2021clipstyler} which uses CLIP score \cite{hessel2021clipscore} as a metric, we use a modified variant of CLIP score called patch-wise CLIP score as a metric to compare image stylization between different benchmarks. We used the same dataset of image-text pairs as in the previous section, but with simplified stylization prompts. What differentiates patch-wise CLIP score from the CLIP score is that CLIP score takes random crops of the whole image, whereas in patch-wise CLIP score, we take 8 random crops per object using the object wise bounding boxes of the image. The object wise bounding boxes are directly extracted from the object wise masks generated by SAM\cite{kirillov2023segment}(Segment Anything Model). Due to this differentiation, we now have a well defined style(text) associated with the random crops that we take from the defined objects bounding boxes. We have evaluated the Patch-wise CLIP scores between the stylized images and the prompts for the following models: LDAST\cite{fu2021languagedriven}, ITStyler\cite{bai2023itstyler}, CLIPStyler\cite{kwon2021clipstyler}, Pix2Pix\cite{isola2016imagetoimage}, Diffstyler\cite{huang2022diffstyler} and MOSAIC. We can observe the effectiveness of Object-wise style transfer of MOSAIC from its CLIP score comparison from Table \ref{Ratings_table}. The rest of the benchmarks tend to stylize the whole image with a mixture of styles instead of Object-wise Style transfer.

\captionsetup{font=large}
\begin{table}
    \centering
    
    \begin{tabular}{|c|c|}
        \hline
            \textbf{Method}      &     \textbf{Rating} $\uparrow$ \\
        \hline
        LDAST\cite{fu2021languagedriven} & 5.7 \\          
        \hline
        ITStyler\cite{bai2023itstyler} & 7.2 \\
        \hline
        CLIP-Styler\cite{kwon2021clipstyler} & 4.6 \\
        \hline
        Pix2Pix\cite{isola2016imagetoimage} & 7.9 \\
        \hline
        Diffstyler\cite{huang2022diffstyler} & 5.5 \\
        \hline
        \textbf{MOSAIC(ours)} &  \textbf{9.1} \\
        \hline

    \end{tabular}

    \Large
    
    \caption{Analysis of User Ratings (ranging from 1 to 10)}
    \label{RESULTS_table}
\end{table}

\captionsetup{font=large}
\begin{table}
    \centering
    
    \begin{tabular}{|c|c|}
        \hline
            \textbf{Method}      &     \textbf{ Patch-wise CLIP score} $\uparrow$\\
        \hline
        LDAST\cite{fu2021languagedriven} & 0.1630 \\          
        \hline
        ITStyler\cite{bai2023itstyler} & 0.2031 \\
        \hline
        CLIP-Styler\cite{kwon2021clipstyler} & 0.1742 \\
        \hline
        Pix2Pix\cite{isola2016imagetoimage} & 0.1934 \\
        \hline
        Diffstyler\cite{huang2022diffstyler} & 0.1697 \\
        \hline
        \textbf{MOSAIC(ours)} &  \textbf{0.2671} \\
        \hline

    \end{tabular}

    \Large
    
    \caption{Patch-wise CLIPScore comparison between state-of-the-art methods and MOSAIC(ours)}
    \label{Ratings_table}
\end{table}

\section{Deployment on Edge Devices}
The following section presents the details on performance of each module in the pipeline. To deploy the pipeline on the edge devices we had to optimize each module individually. Refer to table \ref{deployment_table} for latency's on individual modules and their efficient counter parts.

\begin{table*}[h]
    \centering
    \resizebox{\textwidth}{!}{
    \begin{tabular}{|c|c|c|c|c|}
        \hline
        \multirow{2}{*}{Module} & \multicolumn{2}{c|}{Server(T4 GPU)} & \multicolumn{2}{c|}{Edge} \\ \cline{2-5}

        & Architecture & Latency & Architecture & Latency \\
        \hline
        Segmentation & SAM \cite{kirillov2023segment} & 456 ms & MobileSAM \cite{zhang2023faster} & 12 ms \\ 
        \hline
        Text Encoder & CLIP \cite{cherti2023reproducible} & 286 ms & MoTIS \cite{ren2022leaner} & 98.6 ms \\
        \hline
        Text Segmentation & Transformer Large & 176 ms & Transformer Small & 75.4 ms \\
        \hline
        \hline
        CLIP Score & \multicolumn{2}{c|}{0.2671} & \multicolumn{2}{c|}{0.2245} \\
        \hline
    \end{tabular}
    
    }
    \Large
    
    \caption{Comparision of Latencies of pipeline. For Server deployment we used a single T4 GPU and for edge deployment we benchmarked on Qualcomm SM8450 Snapdragon 8 Gen 1 Processor}
    \label{deployment_table}
\end{table*}

In the context of enhancing the performance and efficiency of our pipeline, several crucial modifications were made to key modules, as outlined below:

Segmentation Module: To achieve improved performance, we replaced the previously employed Segment Anything Model (SAM) \cite{kirillov2023segment} with a more efficient architecture known as MobileSAM \cite{zhang2023faster}. This architectural change resulted in a remarkable reduction in latency from 456ms to 12ms, leading to an impressive 38x speedup.

Text Encoding Module: For encoding text into a unified space, a robust model such as CLIP was initially considered. However, the high inference time of 286ms associated with CLIP posed potential challenges in terms of pipeline latency. To address this concern, we opted to replace CLIP with the text encoder of MoTIS (Mobile Text to Image Search) \cite{ren2022leaner}, which substantially reduced the inference time to 98.6ms, resulting in a nearly 3x speedup.

Text Segmentation Task: While the ChatGPT-based model demonstrated excellent performance and was easily accessible through API calls, we sought to provide an offline version with greater control over output. To accomplish this, custom models, namely Transformer Large (9 Heads) with inference times of 276ms for the large variant and 95.4ms for the smaller variant (3 Heads), were developed.

Stylization Network: Through rigorous testing, it was observed that the Stylization Network exhibited consistent inference times of 30ms. As a result, the overall pipeline's inference time was determined to be 236ms. This is a noteworthy 5x speedup compared to any Diffusion Based Models that typically require 1 second to generate an image, considering the use of 50 sampling steps.

These adjustments demonstrate substantial enhancements to our pipeline's overall efficiency, making it well-suited for various real-world applications.

\section{Limitations}

\begin{figure}[t]
\begin{center}
\includegraphics[width=1.0\linewidth]{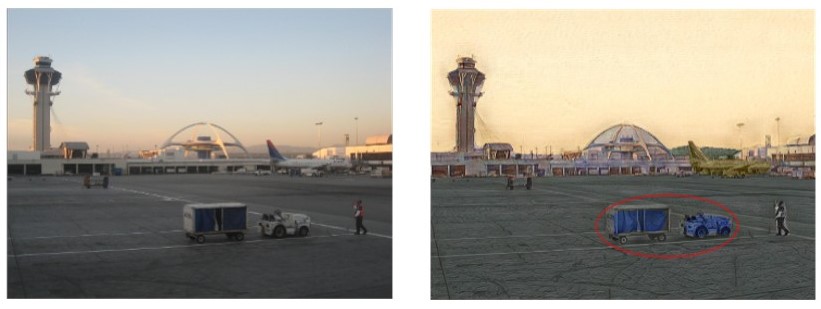}
\end{center}
   \caption{Case of imperfect mask from Image Segmentation Model}   
\label{fig: lim1}
\end{figure}

The main limitation of this model comes with segmentation. The performance of the segmentation model aids in producing more pleasant stylized images. The output sometimes deteriorates due to unpleasant segmentation masks, as shown in Figure \ref{fig: lim1}. 

\begin{figure}[t]
\begin{center}
\includegraphics[width=1.0\linewidth]{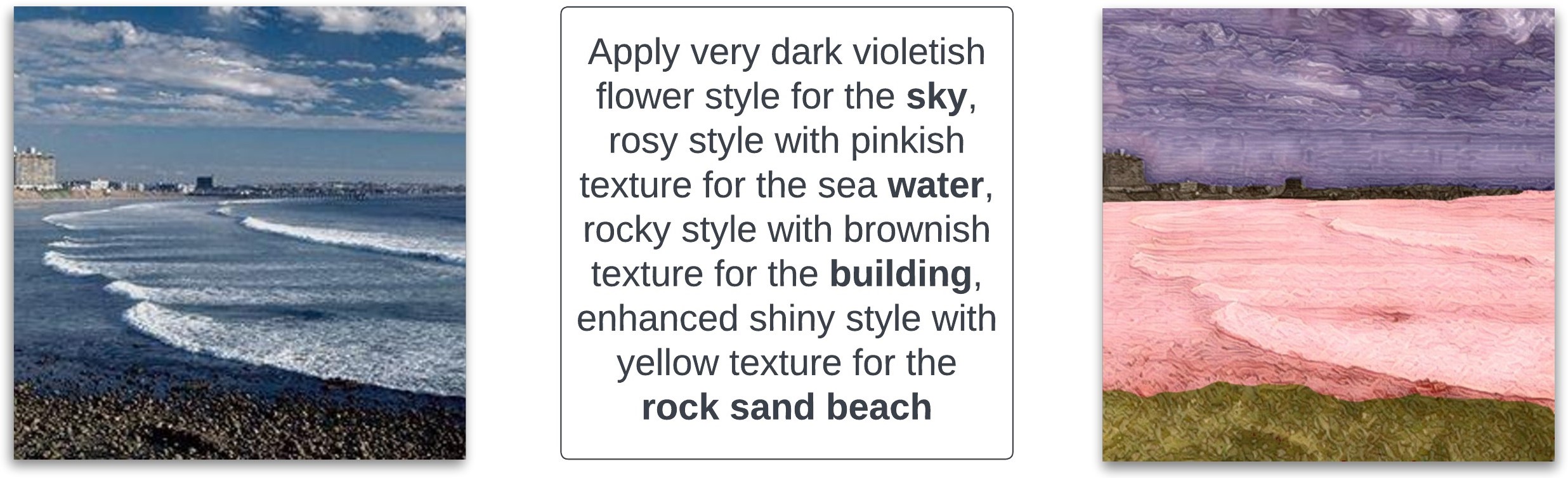}
\end{center}
   \caption{This shows the long prompt even for an Image with fewer objects.}
\label{fig: lim2}
\end{figure}

The other thing is about the long input prompts. This pipeline demands longer prompts for giving better pleasing outputs. This may look clumsy, as shown in Figure \ref{fig: lim2}.

\section{Future Goals}

In the future, we plan to improve the quality of our outputs and streamline our pipeline by integrating multiple modules into a single model using architectures such as GANs\cite{goodfellow2014generative}. Additionally, we aim to enhance the model's ability to stylize individual objects rather than the entire image. The stylization of the whole picture results in many unseen artifacts are being produced due to the mixing of content from the out-of-mask regions. We can achieve more pleasing results by extending this method to do mask-aware style transfer. Due to masks, there was a sudden change in texture or color, sometimes creating an unpleasant look. For this, we can use bilateral grid mapping\cite{Chen2016BilateralGU} and implement a smooth transition. Using bilateral grid mapping to generate the final output gives us aesthetically looking images as they are good at producing photo-realistic images. We also intend to investigate the optimal sequence of blocks through which the input data should flow to achieve the best results. This mask problem can also be addressed using Diffusion models as shown in \cite{Jimnez2023MixtureOD}. Our method can be further extended to future language-based stylization models, so that the stylization would be object wise and well controlled by the user. By performing these modifications, we aim to make further advancements in object-wise stylization using text and contribute to the field's development. 



{\small
\bibliographystyle{ieeetr}
\bibliography{main}
}

\end{document}